\PassOptionsToPackage{no-math}{fontspec}
\documentclass[displayweek]{sid-arxiv}
\usepackage{array}
\usepackage{stfloats}
\usepackage{microtype}
\usepackage{url}
\hypersetup{pageanchor=false,colorlinks=true,urlcolor=black,citecolor=black,linkcolor=black}
\fnbelowfloat
\makeatletter
\renewenvironment{thebibliography}[1]{%
 \section{References}\small\raggedright
 \list{\@arabic\c@enumiv.}{\settowidth\labelwidth{#1.}%
 \leftmargin\labelwidth\advance\leftmargin\labelsep
 \usecounter{enumiv}\setlength\itemsep{4pt}\setlength\parsep{0pt}}%
 \sloppy\clubpenalty4000\widowpenalty4000\sfcode`\.=1000\relax
}{\endlist}
\makeatother
\begin{document}
\pagenumbering{gobble}
\title{Goal-Driven Autonomous Optical Experimentation via a Self-Recursive Hardware--Software Harness}
\author{Tianxing Zhu\textsuperscript{*}, Huacheng Lu\textsuperscript{*}, Ming Liu\textsuperscript{**}, and Malte Hillebrand\textsuperscript{*}\\
\affiliations \textsuperscript{*}Konica Minolta (China) Investment Ltd., Shanghai, China\\
\textsuperscript{**}Giant Testing Equipment Ltd., Shanghai, China}
\date{}
\maketitle
\small

\begin{abstract}
We propose goal-driven self-recursive optical research in which agents use physical evidence to revise the hardware, software, models, and protocols used to pursue a scientific goal. The framework can be instantiated with different stimulus systems and measurement devices. We formalize its updates and derive a relation between physical attainability, command access, and model approximation error. We demonstrate the concept in an optical see-through color-matching case study using an iPad, RayNeo X3 glasses, and a TOP 300 probe coupled to a CAS140D spectroradiometer, both from Instrument Systems. A revised glasses application adds USB-controlled RGB holds and frame acknowledgements, keeping model-generated colors displayed during spectral acquisition. The interface passes 12 live control checks. Spectral feedback reduces one target color difference from 19.756 to 1.211 $\Delta E_{76}^{*}$; all five targets improve in a separate comparison. These measurements demonstrate the concept in one configuration. The resulting software and protocols remain available for subsequent investigations.
\end{abstract}


\section{Introduction}

An optical investigation often requires a new stimulus application, coordinated interfaces for measurement devices, or a calibration method suited to the phenomenon being measured. Developing these capabilities is part of the research. An autonomous system must be able to implement scientific decisions on physical equipment and use the resulting evidence to revise both the experiment and its supporting applications.

Self-driving laboratories combine experimental design, physical execution, and inference in closed loops \cite{r1}. A mobile robotic chemist demonstrated autonomous exploration of photocatalytic formulations \cite{r2}, while scientific agents use language-based reasoning to operate computational and laboratory tools \cite{r3,r4}. We consider how such a system can establish and develop the hardware and software environment needed for a scientific goal.

We introduce goal-driven self-recursive optical research through an integrated hardware and software harness comprising the apparatus, device applications, measurement interfaces, and analysis tools available to agents. The concept can be instantiated in different experimental configurations. To demonstrate it, we develop a case study in which agents create applications for iPad background generation, RayNeo AR glasses rendering, and optical measurement. Scientific planning determines the experiments, and progress monitoring evaluates the evidence against the goal.

Once the harness is established, a result can prompt a different test or a revision to the software, measurement protocol, or analytical method. Agents implement the revision and resume physical testing. We call this process self-recursive because evidence changes the means used to obtain further evidence. The scientific goal persists across these changes, including the joint adaptation of experiments and methods.

As a study and demonstration of the concept, we use optical see-through color compensation, in which display emission combines with transmitted background light at the viewing pupil \cite{r5,r6}. A question about temporal response or spatial variation may require new capabilities in the stimulus and measurement applications. A prediction discrepancy may require a different response model. This case study examines these revisions within one research environment and illustrates how physical attainability, available commands, and model accuracy affect the optical outcome.

\section{Related Work}

\subsection{Autonomous laboratories and scientific agents:} Self-driving laboratories combine learning algorithms with automation and instruments in chemistry and materials science \cite{r1}. ChemOS 2.0 orchestrates heterogeneous experimental and computational components \cite{r7}, and distributed knowledge graphs represent laboratory resources and experimental knowledge \cite{r8}.

Scientific agents extend the reasoning and development work performed in these environments. Coscientist combines language models with search, computation, and laboratory capabilities \cite{r3}. ChemCrow gives language models access to specialized chemistry tools \cite{r4}. Work on scientific facilities examines retrieval and tool augmentation \cite{r9}, while agents that learn instrument operation retain procedural knowledge from experience \cite{r10}. These capabilities support the development of applications and protocols through which agents operate equipment.

A stimulus application determines which commands the system can execute. An acquisition interface determines which physical distinctions it can measure, and a protocol defines the experimental comparison. We treat these as revisable components of the harness, with development directed by the scientific goal and the evidence already acquired.

\subsection{Optical discovery and programmable experimental environments:} AutoSciLab combines autonomous experimentation with interpretable scientific modeling in nanophotonics \cite{r11}. Optical see-through displays offer another setting for optical autonomy: programmable display emission combines with background light in a defined geometry \cite{r5,r6}. The response depends on the stimulus, background, and viewing position, so the investigation requires coordinated control and observation.

The required capabilities depend on the question. Spectral measurements require radiometric observation; spatial measurements require image formation and registration. Settling studies require stimulus sequences and observations that resolve the changing response. Goal-driven self-recursion includes developing the applications, measurement strategies, and models needed as these questions arise.

\begin{figure}[!tb]
\centering
\includegraphics[width=\linewidth,keepaspectratio]{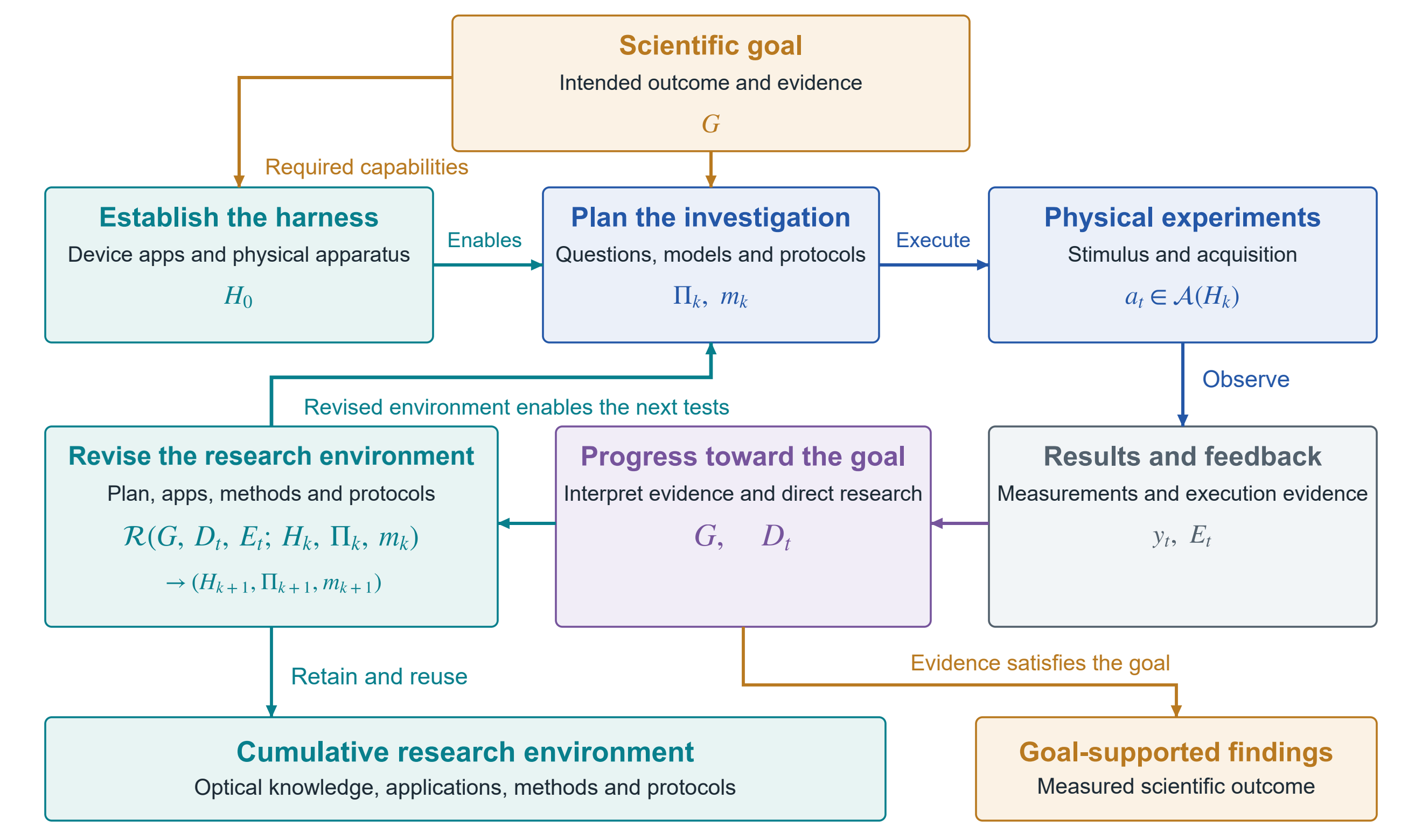}
\caption{Goal-driven self-recursion with the notation of Eqs. (1) and (2). The goal $\mathbf{G}$ directs establishment of $\mathbf{H}_{\mathbf{0}}$, experimental planning, and execution. Optical observations and execution evidence inform the recursive update of the harness, plan, and method. The revised environment enables subsequent tests, while accumulated evidence is evaluated against the same scientific goal.\label{fig:1}}
\end{figure}

\section{Goal-driven self-recursion}

\subsection{A persistent goal and an evolving research environment:} The scientific goal specifies the intended outcome, its application domain, and the evidence used to assess progress. In color compensation, it may be to reproduce a family of target appearances across specified backgrounds and viewing conditions. Individual target matches test parts of that objective. The goal determines which measurements and capability revisions are useful and when the evidence supports the intended outcome.

At research stage k, the harness includes the physical apparatus, device and observation applications, and their connection to scientific computation. The plan organizes the questions to be tested. The method specifies the response models and inferential procedures. Together with the goal and accumulated evidence, they determine a physical action:

\begin{equation}
\begin{gathered}
a_{t} = \pi(G,D_{t};H_{k},\Pi_{k},m_{k}),\quad\quad a_{t}\in\mathcal{A}(H_{k}), \\
y_{t} = \mathcal{O}_{H_{k}}\lbrack F_{phys}(a_{t})\rbrack + \varepsilon_{t}.
\end{gathered}
\end{equation}

Here $G$ is the scientific goal, $D_{t}$ is the accumulated evidence, and $H_{k}$ is the integrated harness. The plan is $\Pi_{k}$ and the method is $m_{k}$. The action set $\mathcal{A}(H_{k})$ contains the tests that the harness can execute. An action can specify a display command, a background condition, and an acquisition configuration. The observation operator describes how the instruments and acquisition applications convert the physical response into evidence.

The apparatus supplies the physical interaction, and the software gives the agents access to it. Adding an application feature or acquisition strategy therefore changes which experiments the system can perform.

\subsection{Evidence-driven revision of the means of investigation:} Experimental evidence updates the research environment and scientific method:

\begin{equation}
\begin{gathered}
(H_{k + 1},\Pi_{k + 1},m_{k + 1}) \\
=\mathcal{R}(G,D_{t},E_{t};H_{k},\Pi_{k},m_{k}).
\end{gathered}
\end{equation}

The evidence $E_{t}$ includes execution feedback and discrepancies that help explain the observation. The revision operator can change the test plan, extend stimulus-control applications, refine acquisition through the available measurement devices, replace a model, or revise the measurement protocol. The updated harness and plan implement the change, and subsequent tests measure its consequences.

The goal determines which revisions to pursue. A new capability may enable an informative test, a model revision may improve the measured optical response, and a plan revision may resolve a remaining scientific question.

\subsection{Goal-relative progress across experimental contexts:} Improving a display setting and revising the method that selects settings are different research actions. For an optical context defined by a target, background, and geometry, the method selects a command available through the harness:

\begin{equation}
\begin{gathered}
u_{H,m}(c) = \underset{u\in\mathcal{U}(H)}{arg\, min}\,\mathcal{l}_{m}({\widehat{F}}_{m}(u,b,g),q), \\
K(H,m;c) = d(F_{phys}(u_{H,m}(c),b,g),q), \\
c = (q,b,g).
\end{gathered}
\end{equation}

The physical outcome $K(H,m;c)$ measures performance in that context. Outcomes across contexts describe how model and capability revisions affect the goal. A distribution $\mu_{G}$ specifies the target and context domain:

\begin{equation}
\begin{gathered}
R_{G}(H,m) = \mathbb{E}_{c \sim \mu_{G}}\lbrack K(H,m;c)\rbrack, \\
\Gamma_{G} = R_{G}(H,m) - R_{G}(H',m').
\end{gathered}
\end{equation}

Positive gain means that the revised harness and method reduce expected mismatch over the goal domain. This measure covers changes to the response method and to available commands. New observation capabilities can also advance the investigation by resolving a scientific question. Progress is assessed using the outcome specified by the goal \cite{r12}.

\begin{figure}[!tb]
\centering
\includegraphics[width=\linewidth,keepaspectratio]{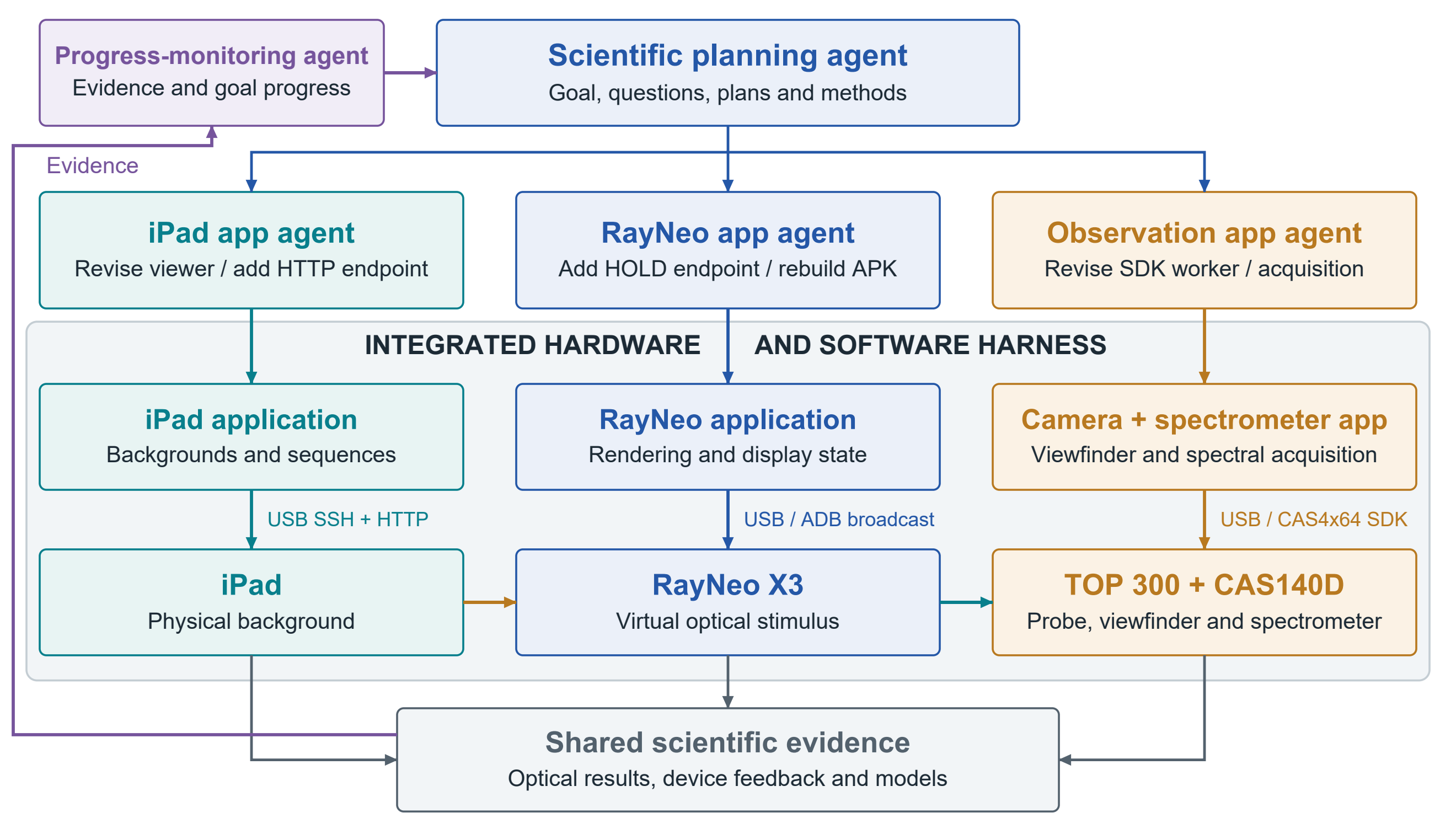}
\caption{Agent responsibilities and implemented interfaces in the case study. The iPad viewer uses HTTP over a USB SSH connection; the glasses application receives USB/ADB commands; spectral acquisition uses the CAS4x64 SDK over USB. The TOP 300 integrates a viewfinder with the fiber-coupled optical probe \cite{r13}. Development examples identify source-level changes such as adding a HOLD command entry point, extending a viewer endpoint, or revising the SDK worker.\label{fig:2}}
\end{figure}

\begin{table*}[!tb]
\centering
\caption{Agent responsibilities in the demonstration platform and the capabilities they establish and revise.\label{tab:1}}
\small
\renewcommand{\arraystretch}{1.16}
\setlength{\tabcolsep}{4pt}
\begin{tabular}{|>{\raggedright\arraybackslash}p{\dimexpr 0.21430\linewidth-2\tabcolsep-1.4\arrayrulewidth\relax}|>{\raggedright\arraybackslash}p{\dimexpr 0.28570\linewidth-2\tabcolsep-1.4\arrayrulewidth\relax}|>{\raggedright\arraybackslash}p{\dimexpr 0.50000\linewidth-2\tabcolsep-1.4\arrayrulewidth\relax}|}
\hline
\textbf{Agent role} & \textbf{Harness contribution} & \textbf{Goal-directed development} \\ \hline
Scientific planning & Experimental questions, models, and test plans & Select informative tests and revise the investigation using new evidence \\ \hline
iPad application development & Background generation and stimulus-state observation & Add background patterns, sequences, or control features required by new questions \\ \hline
RayNeo application development & Glasses rendering and display control & Extend stimulus representation and rendering behavior for subsequent tests \\ \hline
Observation application development & Camera-spectrometer acquisition and analysis interfaces & Relate spatial and spectral observations and develop new acquisition protocols \\ \hline
Progress monitoring & Interpretation of progress against the persistent goal & Identify unresolved objectives and direct testing, analysis, or capability development \\ \hline
\end{tabular}
\end{table*}

\section{Demonstration platform and harness development}

\subsection{Hardware and software integration:} In the developed case study, the harness combines the apparatus and optical geometry with device applications, measurement interfaces, analysis tools, and the agents that develop and use them. It carries a research question through physical execution to a measured response. The stimulus, background, viewing condition, and acquisition settings give that response its experimental context.

In this demonstration platform, the iPad generates the background and the RayNeo X3 supplies display emission. A TOP 300 optical probe and a CAS140D spectroradiometer, both from Instrument Systems, provide optical measurement \cite{r13}. The probe includes an alignment viewfinder and couples the collected light to the spectroradiometer. Device applications generate stimuli, acquisition software records spectral and spatial observations, and analysis tools fit response models and select further tests.

Establishing the harness starts with the stimuli and observations required by the goal. Development agents create or extend the applications, while optical characterization relates the controllable device states to measured responses.

\subsection{Agent responsibilities:} In this case study, scientific planning, application development, and progress monitoring share a goal and the physical evidence from the investigation (Fig. 2). Table 1 assigns their responsibilities across the device and instrument applications.

The iPad development agent builds background-generation features, and the RayNeo development agent builds rendering and stimulus control. The observation development agent builds the camera-spectrometer application and its acquisition interfaces.

The progress-monitoring agent evaluates results against the goal and test plan. It identifies unresolved objectives and directs the next measurement, analysis, or application revision. This role includes deciding when another measurement is useful and when the investigation first needs a new capability.

\subsection{Illustration of recursive development:} As a possible extension of the case study, consider an investigation of target appearance across changing backgrounds. An initial harness generates static backgrounds, renders virtual colors, and acquires spectra for a response model. Suppose new observations depend on the preceding stimulus. The next comparison must distinguish optical memory from a static response discrepancy and may require new background and rendering sequences with time-resolved acquisition.

\begin{figure*}[!tb]
\centering
\includegraphics[width=\linewidth,keepaspectratio]{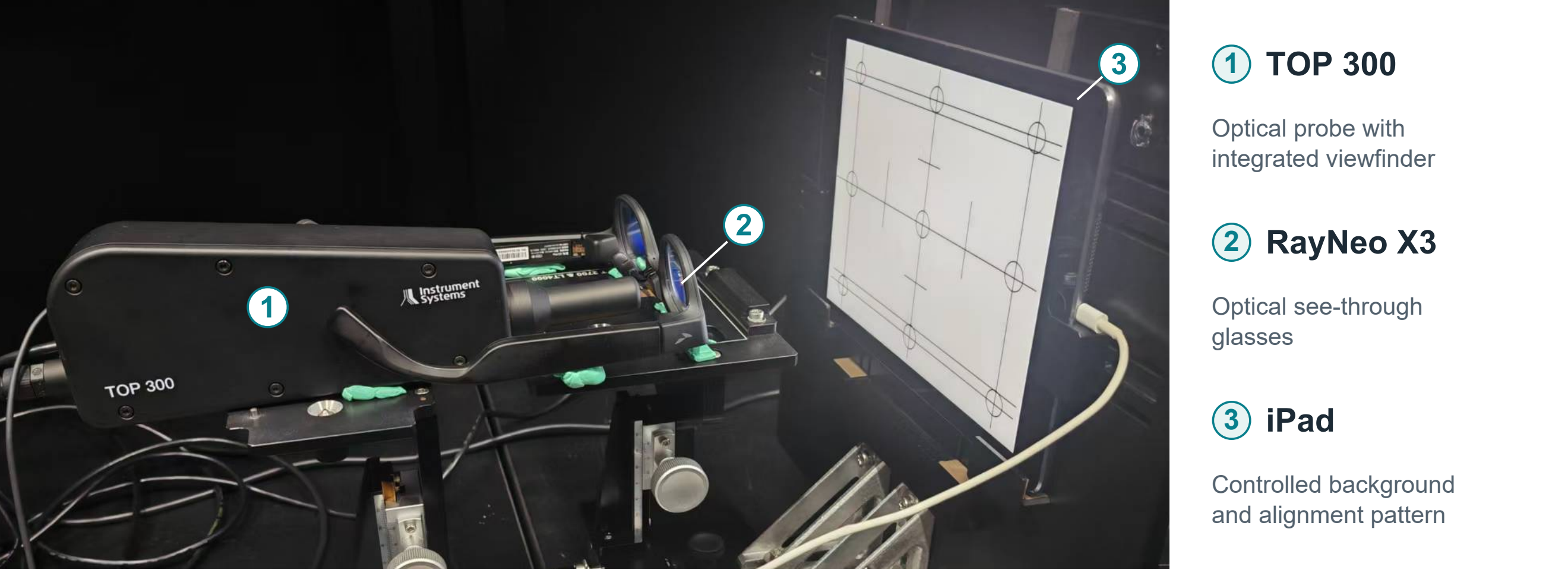}
\caption{Photograph of the demonstration platform. (1) TOP 300 optical probe, (2) mounted RayNeo X3 glasses, and (3) iPad displaying an alignment pattern. Adjustable translation stages position the glasses between the background display and the probe viewing pupil. The probe is fiber-coupled to the CAS140D spectroradiometer.\label{fig:3}}
\end{figure*}

The planning agent specifies that comparison. Device development agents extend the iPad and glasses applications to generate the sequences, and the observation development agent extends acquisition to resolve the response. These tests inform the response model and measurement protocol. Progress monitoring evaluates the result against the original appearance objective and selects the next investigation.

The progress-monitoring agent checks revisions against the measurement relationships they must preserve. A measurement transaction records the requested stimulus, device acknowledgement, acquisition settings, and spectrum. Invariant checks cover stimulus identity, calibration and units, display stability, and command-observation correspondence. A revision capsule records the scientific motivation, source or build fingerprint, and reference checks. Known stimuli and retained reference observations allow comparisons across application versions. In Section 6.2, command validation and framebuffer checks precede spectral acquisition with the revised interface.

\subsection{Physical optical arrangement:} The iPad background and RayNeo X3 emission enter a common viewing pupil (Fig. 3). The TOP 300 probe collects the light and guides it through a fiber to the CAS140D spectroradiometer. Its integrated viewfinder camera supports alignment \cite{r13}. Device applications control the stimuli, and acquisition software records the spectral response used for the colorimetric tests in Section 6.

\section{Optical formulation}

\subsection{Additive light formation and the attainable set:} For an optical see-through display, a first-order spectral description at a fixed viewing geometry is

\begin{equation}
\begin{gathered}
S_{obs}(\lambda;u,b,g) = S_{AR}(\lambda;u,g) \\
 + \tau(\lambda,g)S_{bg}(\lambda;b,g) + \epsilon(\lambda).
\end{gathered}
\end{equation}

Display emission and transmitted background light combine at the observation plane, with spectral transmittance weighting the background contribution. Integration against the CIE color-matching functions gives tristimulus values, which are transformed to CIELAB using a specified reference white \cite{r14}. We denote the physical color response by $F_{phys}(u,b,g)$.

The physical command domain defines the attainable color set for a background and geometry. The harness exposes a subset of those commands, giving two target floors:

\begin{equation}
\begin{gathered}
\mathcal{G}_{b,g} = \{ F_{phys}(u,b,g):u \in \mathcal{U}_{phys}\}, \\
\mathcal{U}(H) \subseteq \mathcal{U}_{phys}, \\
L_{phys}^{*}(c) = \min_{u \in \mathcal{U}_{phys}}d(F_{phys}(u,b,g),q), \\
L_{H}^{*}(c) = \min_{u\in\mathcal{U}(H)}d(F_{phys}(u,b,g),q).
\end{gathered}
\end{equation}

Display commands add light to a background contribution set by optical transmission. The attainable set therefore depends on the background and geometry. An idealized linear additive system translates the response set in XYZ. Nonlinear device response, wavelength-dependent transmission, and conversion to CIELAB produce a more general geometry.

Extending a stimulus application can expose additional physically supported commands and lower the executable target floor. Extending acquisition can make a response observable. The apparatus defines the physical interactions; the harness determines which of them agents can use and measure.

\subsection{Error bound for command access and model approximation:} Consider a method that globally minimizes predicted target distance over the command domain exposed by the harness. At a fixed context, suppose its uniform approximation error satisfies

\begin{equation}
\begin{gathered}
\sup_{u\in\mathcal{U}(H)}d({\widehat{F}}_{m}(u,b,g),F_{phys}(u,b,g)) \\
 \leq \epsilon_{H,m}(c).
\end{gathered}
\end{equation}

The harness access gap is the difference between the executable and physical target floors. The selected command then satisfies

\begin{equation}
\begin{gathered}
\delta_{H}(c) = L_{H}^{*}(c) - L_{phys}^{*}(c) \geq 0, \\
L_{phys}^{*}(c) \leq K(H,m;c) \\
 \leq L_{phys}^{*}(c) + \delta_{H}(c) + 2\epsilon_{H,m}(c).
\end{gathered}
\end{equation}

Apply the triangle inequality to the selected physical response, its prediction, and the target. Optimality in the predicted problem bounds the predicted distance by that of an optimal executable command. A second application of the triangle inequality gives the executable target floor plus twice the model approximation error. Substitution of the access-gap definition gives the upper bound in Eq. (8). Physical optimality over the full command domain gives the lower bound.

\begin{figure*}[!tb]
\centering
\includegraphics[width=\linewidth,keepaspectratio]{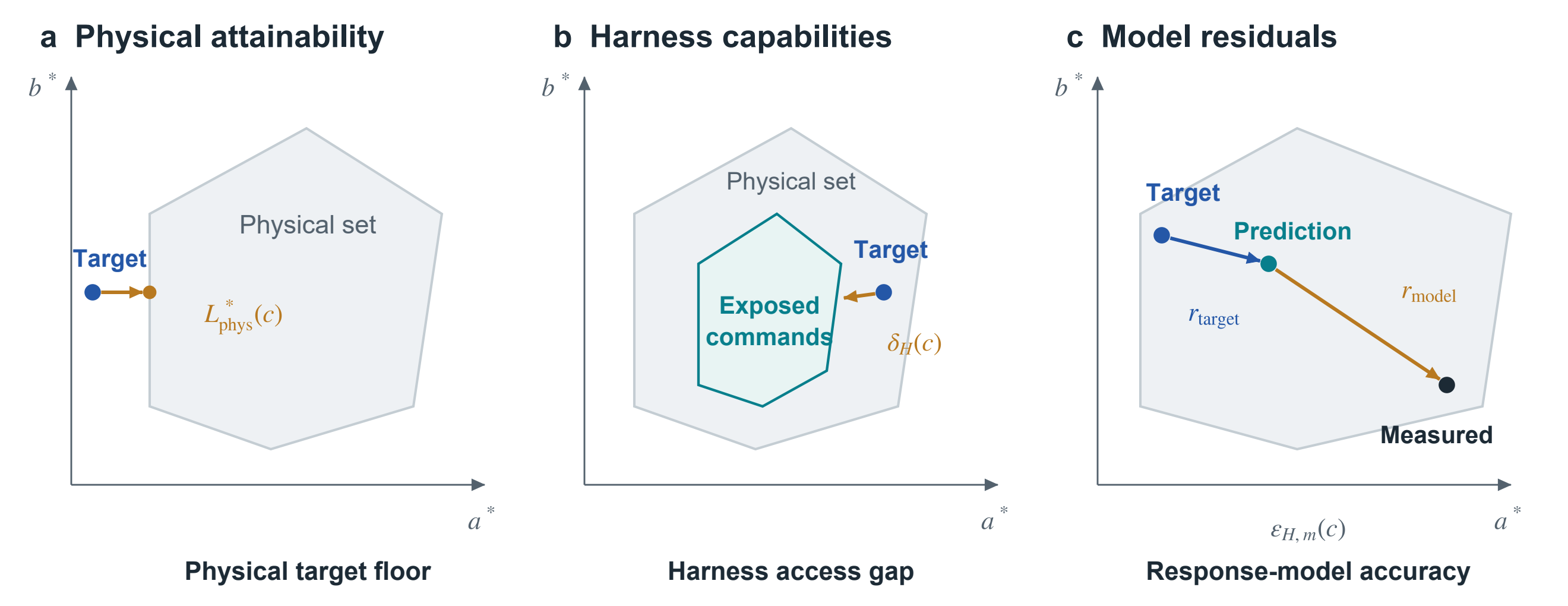}
\caption{Geometric interpretation of Eqs. (6)-(9) in schematic CIELAB slices, with $\mathbf{a}^{\mathbf{*}}$ and $\mathbf{b}^{\mathbf{*}}$ axes at fixed $\mathbf{L}^{\mathbf{*}}$. (a) The minimum target distance is the physical target floor, $\mathbf{L}_{\mathbf{phys}}^{\mathbf{*}}\mathbf{(}\mathbf{c}\mathbf{)}$. (b) The restricted command set introduces the harness access gap, $\boldsymbol{\delta}_{\mathbf{H}}\mathbf{(}\mathbf{c}\mathbf{)}$. (c) The two residual vectors are $\mathbf{r}_{\mathbf{target}}$ from target to prediction and $\mathbf{r}_{\mathbf{model}}$ from prediction to measurement. Equation (7) defines the uniform model approximation quantity, $\boldsymbol{\epsilon}_{\mathbf{H}\mathbf{,}\mathbf{m}}\mathbf{(}\mathbf{c}\mathbf{)}$, over the executable domain.\label{fig:4}}
\end{figure*}

Equation (8) separates the effects of the physical configuration, command access, and response model. Changing the apparatus or geometry can change the attainable set. Extending command access can reduce the harness gap, and improving the model can reduce approximation error. The gap is zero when the application exposes the full physical command set. Observation capabilities affect the evidence used to learn and apply the response model.

CIELAB Euclidean distance supplies the metric structure for this derivation. Equation (7) defines a uniform error over the executable command domain; empirical prediction scores describe sampled responses.

\subsection{Residuals and experimental design:} For an observed optical response, the target error admits the vector decomposition

\begin{equation}
\begin{gathered}
r_{target} = {\widehat{F}}_{m}(u,b,g) - q, \\
r_{model} = y - {\widehat{F}}_{m}(u,b,g), \\
y - q = r_{target} + r_{model}.
\end{gathered}
\end{equation}

The first residual is the predicted displacement from the target. The second is the difference between the measured and predicted response. A small predicted target residual with a large measured mismatch directs attention to the response model. Dependence on command, background, or target region suggests variables and representations to test.

Repeated references under fixed geometry characterize variation around a stimulus. Changing the background while holding the target probes background dependence; neighboring commands near a boundary probe local response and control authority. Comparisons across model families test which representation predicts the relevant optical behavior. When a comparison requires an unavailable stimulus or observation, the agents must revise the application or acquisition protocol before performing it.

\section{Optical color-matching case study}

\subsection{Color matching and characterization:} The case study seeks to match a desired color through the glasses while background light enters the viewing path. It uses measured spectra to update the display setting and to revise the response model that selects settings for later tests.

We calculate color from the measured spectrum and report Euclidean distance in CIELAB as $\Delta E_{76}^{*}$, using reference white XYZ = (95.047, 100, 108.883) \cite{r14}. Smaller values indicate closer colorimetric matches. Presentation and viewing conditions determine the visual interpretation of these differences \cite{r15}. The characterization data contain 2,170 stimulus conditions and 356 additional control measurements for learning the response to display settings and backgrounds.

The corpus combines 320 direct iPad calibrations (64 patterns at five brightness settings) with 1,850 combined-platform conditions: 320 transmission, 960 white-background/display, 378 colored-background/display, and 192 background-occluded display-only measurements. The host acquired the combined conditions and all 356 controls through $H_{1}$’s acknowledged HOLD/BLACK interface; the 960-condition block, for example, crosses five iPad brightness settings, three glasses brightness settings, and 64 patterns.

\subsection{Harness establishment and a revised stimulus protocol:} The color-matching loop requires model-generated RGB settings to remain displayed throughout an exposure-dependent spectrometer acquisition. The glasses application accepts arbitrary RGB HOLD commands through a USB/ADB entry point. It retains the requested color and reports the command token, session, eye selection, and draw acknowledgements. The host coordinates acquisition with this held state and stores its readback with the spectrum.

\begin{figure*}[!tb]
\centering
\includegraphics[width=\linewidth,keepaspectratio]{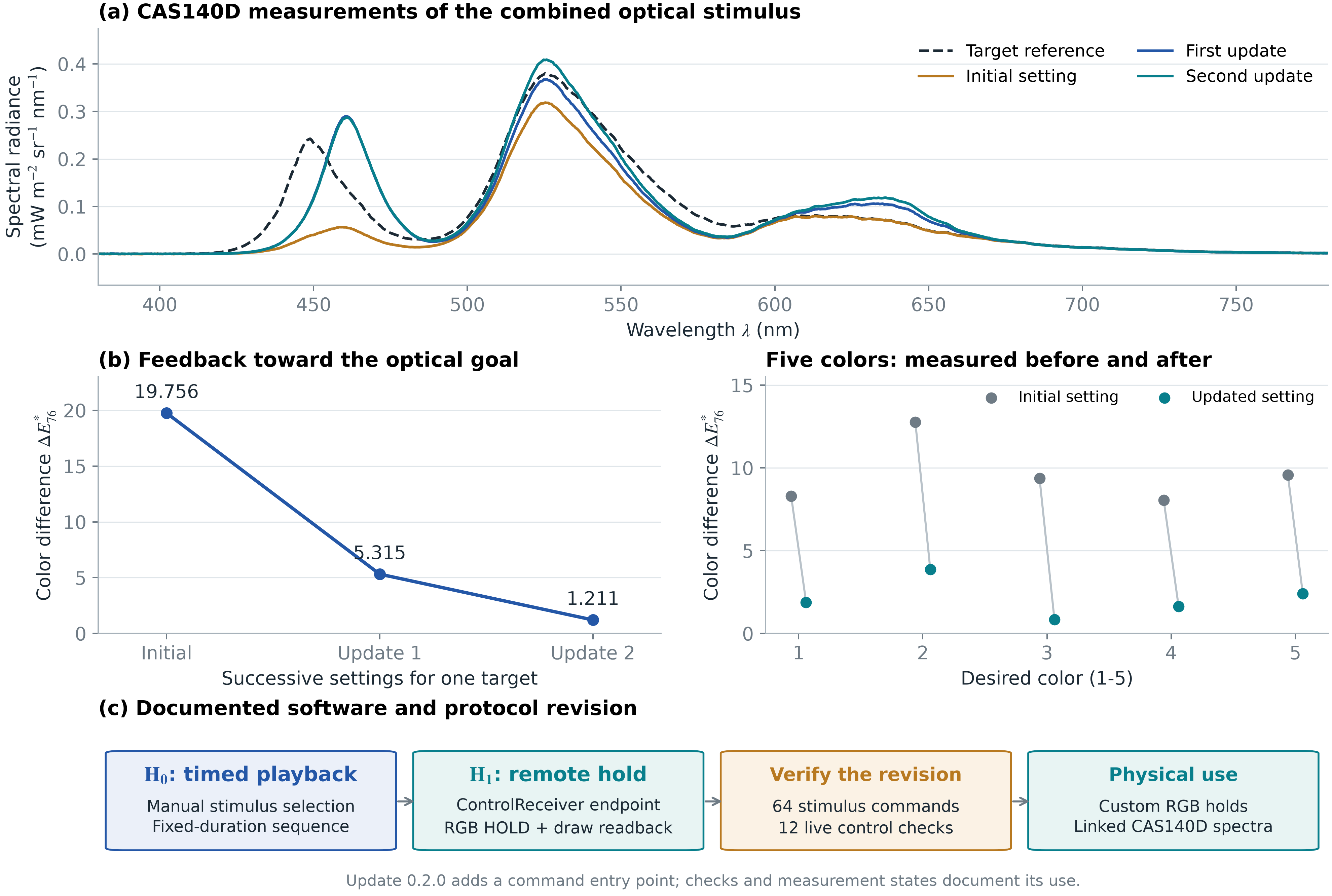}
\caption{Physical measurements and a documented harness revision. (a) Four calibrated combined-light spectra: the target reference under the neutral background and the initial, first-updated, and second-updated settings under the changed background. The original W/(cm\textsuperscript{2} sr nm) calibration is converted to mW/(m\textsuperscript{2} sr nm). (b) Left: the measured three-setting trajectory for one target; right: independent before-and-after measurements for five desired colors. Smaller $\boldsymbol{\Delta}\mathbf{E}_{\mathbf{76}}^{\mathbf{*}}$ indicates a closer color match. (c) The command-interface revision, its device checks, and its subsequent use for spectral acquisition.\label{fig:5}}
\end{figure*}

The transition from $H_{0}$ to $H_{1}$ adds a command receiver that is absent from the baseline binary and present in the updated application. Validation covers all 64 named stimulus commands and passes 12 live device-control tests. Three framebuffer checks each sample 384 pixels and find zero RGB mismatches. Each spectrum in the feedback example retains a custom RGB HOLD state with both software views acknowledged.

An acknowledged hold lets a fitted model request a color and measure its physical response over the acquisition interval. The retained revision record includes the added receiver, source fingerprints, device-check results, and command states stored with the measurements (Fig. 5c).

At fixed eye, patch geometry, and brightness, the revised interface expands catalog-based playback to the full digital RGB domain, $\mathcal{U}(H_{0}\mathcal{) \subset U(}H_{1}) = \{ 0,\ldots,255\}^{3} = \mathcal{U}_{phys}$. Equation (8) then gives $\delta_{H_{1}}(c) = 0$ for RGB command access. Inverse-model outputs are rounded to the device grid and held for measurement.

\subsection{Feedback-guided color matching:} For one fixed target, the initial display setting gives a color difference of 19.756 $\Delta E_{76}^{*}$. Two feedback-guided updates reduce it to 5.315 and 1.211 $\Delta E_{76}^{*}$, a reduction of 93.9\% (Fig. 5a,b).

The reference and updated spectra have blue peaks near 449 and 460-461 nm, consistent with different background and virtual-display primaries. Their combination gives a near-colorimetric match of 1.211 $\Delta E_{76}^{*}$ through metameric compensation in the specified colorimetric system \cite{r16}.

In a separate five-target comparison, the median color difference falls from 9.377 to 1.882 $\Delta E_{76}^{*}$. All five targets improve.

\subsection{Response-model revision:} The response model predicts color from the display setting and background, then supports inverse selection of a setting for a desired color. Revising this model changes the rule used to select settings in subsequent tests.

We compare three response families on the expanded calibration set: a diagonal XYZ affine model fitted in XYZ, the same parameterization fitted by minimizing fixed-reference CIELAB residuals, and an extension with bounded channel-wise quadratic terms fitted with the CIELAB objective. Each family produces 68 predictions with whole target-color groups excluded from its fit. Prediction RMSE values are 1.842, 1.363, and 1.444 $\Delta E_{76}^{*}$, respectively. We select the CIELAB-objective affine model, refit it on the expanded evidence, and acquire new spectra for the settings it generates.

The previous and updated models each select settings for three target colors under three backgrounds, giving nine matched tests and 18 spectra. The updated model gives closer matches in seven tests and larger differences in two. The previous model has a median physical difference of 1.342 $\Delta E_{76}^{*}$. The median decrease across paired tests is 0.197 $\Delta E_{76}^{*}$. This comparison measures calibration refinement across the matched contexts; Section 6.3 measures feedback adjustment of the display command.

\section{Discussion}

Stimulus and acquisition applications are research outputs that determine which comparisons agents can execute and which responses they can observe. Equation (8) separates their optical effects: apparatus changes affect physical attainability, application changes affect command access, and model revisions affect prediction error. A measured discrepancy can therefore direct attention to a specific part of the research environment.

\begin{table*}[!tb]
\centering
\caption{Optical evidence and its role in goal-driven recursion. Color difference is measured in $\boldsymbol{\Delta}\mathbf{E}_{\mathbf{76}}^{\mathbf{*}}$; smaller values indicate a closer match to the desired color.\label{tab:2}}
\small
\renewcommand{\arraystretch}{1.16}
\setlength{\tabcolsep}{4pt}
\begin{tabular}{|>{\raggedright\arraybackslash}p{\dimexpr 0.21430\linewidth-2\tabcolsep-1.4\arrayrulewidth\relax}|>{\raggedright\arraybackslash}p{\dimexpr 0.28570\linewidth-2\tabcolsep-1.4\arrayrulewidth\relax}|>{\raggedright\arraybackslash}p{\dimexpr 0.50000\linewidth-2\tabcolsep-1.4\arrayrulewidth\relax}|}
\hline
\textbf{Scientific question} & \textbf{Measured outcome} & \textbf{Role in the research process} \\ \hline
Can a new command interface enable the optical test? & Arbitrary RGB holds, 64 checked commands, and 12 passed live control tests & Software and protocol development make model-generated settings executable and observable. \\ \hline
Can feedback improve one displayed color? & Color difference $19.756 \rightarrow 1.211$ after two updates; 93.9\% reduction & An observation changes the next physical action while the target remains fixed. \\ \hline
Can the approach support several desired colors? & For five colors, median difference $9.377 \rightarrow 1.882$; all five improve & A shared control and measurement environment supports a family of optical targets. \\ \hline
Can a revised model improve later tests? & The updated model gives closer matches in seven of nine new physical tests. & Evidence changes the reusable relationship that generates subsequent display settings. \\ \hline
\end{tabular}
\end{table*}

A persistent goal allows the investigation to revise its plan, applications, and response models without changing its intended outcome. Progress monitoring evaluates those revisions against unresolved parts of the objective and the evidence needed to resolve them.

Discrepancies can expose assumptions in the optical model or the harness. A structured residual may reveal background dependence. A mismatch between requested and observed stimuli can identify a device state that needs readback. Dependence on acquisition conditions can require a different measurement or analysis protocol. Agents turn these observations into hypotheses and revisions, then acquire the physical comparisons needed to assess them.

Further work can examine transfer between goals, coordination among development agents, and the choice between acquiring another measurement and developing a new capability.

\section{Conclusion}

Goal-driven self-recursive optical research uses physical evidence to revise a goal-specific hardware and software harness. The reported optical see-through color-matching case study demonstrates expanded command access, spectral feedback, and response-model refinement. The optical formulation distinguishes physical attainability, command access, and model approximation error. This configuration illustrates a research structure that can be instantiated with other stimulus systems and measurement devices. The resulting applications and protocols remain available for subsequent investigations.

\section{Code availability}

The case-study software is available in the SpectralColorLab repository: \url{https://github.com/TZHU64/SpectralColorLab}.

\balance

\end{document}